# Constructing Time Series Shape Association Measures: Minkowski Distance and Data Standardization

Ildar Batyrshin
Research Program of Applied Mathematics and Computations
Mexican Petroleum Institute
Mexico D.F., Mexico
E-mail: batyr1@gmail.com

*Abstract*— It is surprising that last two decades many works in time series data mining and clustering were concerned with measures of similarity of time series but not with measures of association that can be used for measuring possible direct and inverse relationships between time series. Inverse relationships can exist between dynamics of prices and sell volumes, between growth patterns of competitive companies, between well production data in oilfields, between wind velocity and air pollution concentration etc. The paper develops a theoretical basis for analysis and construction of time series shape association measures. Starting from the axioms of time series shape association measures it studies the methods of construction of measures satisfying these axioms. Several general methods of construction of such measures suitable for measuring time series shape similarity and shape association are proposed. Time series shape association measures based on Minkowski distance and data standardization methods are considered. The cosine similarity and the Pearson's correlation coefficient are obtained as particular cases of the proposed general methods that can be used also for construction of new association measures in data analysis.

I. INTRODUCTION

The last two decades many works in time series data mining and time series clustering were concerned with measures of similarity of time series [1,2,10-13,16,24,28-30]. These measures are used in indexing and similarity search in time series data bases, in time series clustering, in the search of similarity patterns in time series. But in many applications of time series analysis it is desirable to measure not only similarity between time series but also possible inverse relationships between them. Such relationships can exist between dynamics of prices and sell volumes, between growth patterns of competitive companies, between well production data in reservoirs, between wind velocity and air pollution concentration etc. [3, 5, 6, 8, 17]. An analysis of direct and inverse relationships between variables mainly based on Pearson's correlation coefficient is widely used in almost all applications of statistics: in medicine, biology, economics, physics, social sciences etc. And it is surprising that there is not so much works devoted to time series shape association measures that can be used for measuring possible direct and inverse relationships between time series.

Recently time series shape association measure based on moving approximation transform (MAT) has been introduced and studied in [6]. This measure and its modifications have been used for analysis of possible communications between wells in reservoirs [5] and for analysis associations between atmospheric pollutants and meteorological variables [3]. In [9] a system of axioms defining time series shape association measures has been proposed.

This paper develops a theoretical basis for analysis and construction of time series shape association measures. Starting from the axioms of time series shape association measures it studies the methods of construction of measures satisfying these axioms. Several general methods of construction of such measures suitable for measuring time series shape similarity and shape association are proposed. Time series shape association measures based on Minkowski distance and data standardization methods are considered. The cosine similarity and the Pearson's correlation coefficient are obtained as partial cases of the proposed general methods. Performance evaluation of some of proposed measures of association on a benchmark example of time series is presented.

The paper has the following structure. In Section 2 the definitions of time series shape proximity measures including shape similarity, shape dissimilarity and shape association measures are considered. The relationships between these measures and general methods of construction of shape association measures are presented in Section 3. Section 4 describes the methods of construction of shape association measures based on Minkowski distance. Section 5 discusses the methods of time series standardization and their properties. Section 6 considers some examples of shape association measures obtained by proposed methods. The results of the analysis of the performance of some proposed association measures are presented in Section 7. Discussions and conclusions are given in Section 8.

II. TIME SERIES SHAPE PROXIMITY MEASURES

A time series of length $n$, ($n>1$), is a sequence of $n$ real values $x = (x_1,\ldots,x_n)$. Below we will consider the properties formulated for all time series $x,y$ of the same length $n$. Suppose

$p,q$ are real values. Denote $x+y = (x_1+y_1, ..., x_n+y_n)$, $px+q = (px_1+q, ..., px_n+q)$. A real valued function $P(x,y)$ satisfying the **symmetry** property:

$P(x,y) = P(y,x)$,

will be called a **proximity measure**.

Denote $q_{(n)}$ a constant time series of length $n$ with all elements equal to $q$. We will write $x = const$ if $x = q_{(n)}$ for some $q$, and $x \neq const$ if $x_i \neq x_j$ for some $i \neq j$ from $\{1,...,n\}$. From definitions above it follows: $px+q = px+q_{(n)}$.

A proximity measure will be referred to as:

1) a *dissimilarity measure*, denoted by $D$, if $D(x,y) \geq 0$ and $D(x,x) = 0$. $D$ will be called a **normalized** dissimilarity measure if it takes values in $[0,1]$.

2) a *similarity measure*, denoted by $S$, if it takes values in $[0,1]$ and satisfies:

$S(x,x) = 1$;     (reflexivity)

3) an **association measure**, denoted by $A$, if it takes values in $[-1,1]$ and satisfies the properties:

$A(x,x) = 1$,     (reflexivity)

$A(-x,x) = -1$,  $x \neq const$,     (inverse reflexivity)

$A(-x,y) = -A(x,y)$, $x,y \neq const$.     (inverse relationship)

A proximity function will be called a **shape proximity measure (SPM)** if it satisfies:

$P(x+q,y) = P(x,y)$.     (translation invariance)

Note that a translation $x+q$ of any time series $x$ does not change its shape because all values of $x$ are moved together up or down at the same value $q$. For this reason, the translation invariance is considered as a **necessary requirement** on the shape proximity measures. Translation invariant dissimilarity, similarity and association measures will be referred to as a shape dissimilarity measure (SDM), shape similarity measure (SSM) and shape association measure (SAM), correspondingly.

A shape proximity measure will be called **scale invariant** if it satisfies the property:

$P(px,y) = P(x,y)$,  if $p > 0$.     (scale invariance)

Unlike the translation, a multiplication of all elements of a time series $x$ on a positive constant $p$ changes proportionally the shape of a non-constant time series. Nevertheless, the scale invariance is very useful for scaling time series values (for example, for changing the units of time series from dollars to thousands of dollars), for standardization of time series values measured in different scales etc. For this reason scale invariance will be considered as a **desirable property** of SPM. Discussion of the possible requirements on time series shape similarity and shape association measures can be found in [9].

From the definition of a scale invariant SPM it follows that for all real values $q_1, q_2, p_1, p_2$ such that $p_1, p_2 > 0$ it is fulfilled:

$P(p_1x+q_1, p_2y+q_2) = P(x,y)$.

It is clear that for all real $q_1, q_2, p_1, p_2$ such that $p_1, p_2 \neq 0$ a scale invariant shape association measure satisfies the property:

$A(p_1x+q_1, p_2y+q_2) = sign(p_1) \cdot sign(p_2) \cdot A(x,y)$,

where $sign(x)=1$, if $x>0$; $sign(x)=0$, if $x=0$; $sign(x)=-1$, if $x<0$.

It is clear that a sample Pearson correlation coefficient:

$$corr(x,y) = \frac{\sum_{i=1}^{n}(x_i - \bar{x})(y_i - \bar{y})}{\sqrt{\sum_{i=1}^{n}(x_i - \bar{x})^2 \cdot \sum_{i=1}^{n}(y_i - \bar{y})^2}}$$

is a translation and scale invariant time series shape association measure. The correlation coefficient is used for measuring possible direct and inverse relationships between variables. It is considered as a measure of the strength of linear relationship between variables but it is not relevant for measuring possible associations between variables in general case [4] and for measuring associations between time series shapes [5].

In the definition of correlation coefficient it is supposed that $x,y \neq const$, otherwise we obtain zero in denominator. Generally, below we will suppose that $x \neq const$ if it does not mentioned the opposite. Note that constant time series in some sense similar to zero vector in vector space. For constant time series defined by arbitrary real values $q, r$ and for any time series $y$ of length $n$ the shape similarity measure $S$ satisfies the following properties:

$S(x,q_{(n)}) = S(x,r_{(n)})$,     $S(q_{(n)},r_{(n)}) = 1$.

III. CONSTRUCTION OF SHAPE ASSOCIATION MEASURES

Suppose $D$ is a shape dissimilarity measure and $U$ is a strictly decreasing nonnegative function such that $U(0) = 1$. It is clear that the function

$S_D(x,y) = U(D(x,y))$,

is a shape similarity measure and it is scale invariant if $D$ is scale invariant. For example, we can use one of the following definitions:

$S_D(x,y) = \dfrac{K}{D(x,y)+K}$, $K > 0$,

$S_D(x,y) = \dfrac{1}{e^{D(x,y)}}$.

If it exists some positive constant $H$ such that $H \geq D(x,y)$ for all $x,y$ from a considered set of time series, and $W$ is a strictly increasing function such that $W(0) = 0$, $W(H) \leq 1$, then a shape similarity function can be defined as follows:

$S_D(x,y) = 1 - W(D(x,y))$.

Similarly, we can define a shape dissimilarity measure $D$ by means of shape similarity measure $S$ and a strictly decreasing function $V$ such that $V(1) = 0$:

$D_S(x,y) = V(S(x,y))$.

For example, we can define:

$D_S(x,y)=K(1-S(x,y))$.

The relationships between SSM and normalized SDM can be very simple:

$S_D(x,y)=1-D(x,y)$, $\quad D_S(x,y)=1-S(x,y)$.

Note that the relationships

$S_D(x,y) > S_D(v,w) \quad \text{iff} \quad D(x,y) < D(v,w)$

will not depend on the selection of the specific functions $U, V$ and constants $K, Q$ satisfying to the considered above properties. Hence the result of the clustering of time series based on the similarity measures $S_D$ and $D_S$ will not depend on the selection of $U, V, K, Q$ if the clustering algorithm is invariant under monotone transformations of proximity values [7]. Further, dissimilarity measures defined by Minkowski distance will be considered.

Below we study possible relationships between shape similarity and shape association measures. A time series $-x$ will be called a **reflection of $x$.** Consider the following properties of shape similarity measures $S$ formulated for all time series $x$ ($x \neq const$) of length $n$:

$S(-x,y) = S(x,y)$, $\quad$ (reflection invariance)

$S(x,y) + S(-x,y) = 1$, $\quad$ (complement of reflections)

$S(-x,y) = 1 - S(x,y)$, $\quad$ (complement of reflections)

$S(-x,y) = S(x,-y)$. $\quad$ (sign permutation)

$S(-x,-y) = S(x,y)$. $\quad$ (sign cancellation)

$S(-x,x) = 1$. $\quad$ (similarity of reflections)

$S(-x,x) < 1$. $\quad$ (weak similarity of reflections)

$S(-x,x) = 0$. $\quad$ (non-similarity of reflections)

Note that the sign permutation and sign cancellation are equivalent properties. Indeed, from the sign permutation we have: $S(-x,-y)=S(x,-(-y))=S(x,y)$. Similarly, from the sign cancellation it follows: $S(-x,y)= S(-(-x),-y) = S(x,-y)$.

It is clear that:

1) from the reflection invariance and the reflexivity it follows a similarity of reflections;

2) from the complement of reflections and the reflexivity it follows a non-similarity of reflections.

From these two conclusions we immediately obtain that the reflection invariance and the complement of reflections are incompatible properties for SSM. Also it is easy to see that:

3) the sign permutation and the sign cancellation follow

a) from the reflection invariance and the commutativity, or

b) from the complement of reflections and the commutativity;

4) the weak similarity of reflections follows from the non-similarity of reflections.

It is also clear that if $A$ is a shape association measure then a function

$S_A(x,y) = abs(A(x,y))$,

is a reflection invariant shape similarity measure. $S_A$ is scale invariant if $A$ is scale invariant.

The shape similarity measure $S_A$ evaluates a strength of association between $x$ and $y$, independently of its sign. Such similarity measure is used in [6] in clustering of time series with high absolute value of positive or negative association. As a shape association measure it was used a local trend association measure. The property of a reflection invariance of SSM $S$ can be considered as rational if we want to interpret $S$ as a strength of possible (positive or negative, direct or inverse) relationship between time series. A time series shape similarity measure $S$ will be referred to as an **indirect** shape similarity measure if it is reflection invariant, otherwise, it will be referred to as a **direct** shape similarity measure.

Below we propose two general methods of construction of shape association measures for non-constant time series.

***Theorem* 1.** *Suppose $S$ is a shape similarity measure satisfying the properties of sign permutation and weak similarity of reflections then the function $A_S(x,y)$ defined for all $x,y \neq const$, as follows:*

$$A_S(x,y) = \begin{cases} S(x,y), & \text{if} \quad S(x,y) > S(x,-y) \\ -S(x,-y), & \text{if} \quad S(x,y) < S(x,-y) \\ 0, & \text{otherwise} \end{cases}$$

*is a shape association measure. It is scale invariant if $S$ is scale invariant.*

***Theorem* 2.** *Suppose $S$ is a shape similarity measure satisfying the properties of sign permutation and the non-similarity of reflections then the function $A_S(x,y)$ defined for all $x,y \neq const$ by:*

$A_S(x,y) = S(x,y) - S(x,-y)$,

*is a shape association measure. $A_S$ is scale invariant if $S$ is scale invariant.*

***Corollary* 1.** *Suppose $S$ is a shape similarity measure satisfying the complement of reflections property then the function $A_S(x,y)$ defined for all $x,y \neq const$ by:*

$A_S(x,y) = 2S(x,y) - 1$,

*is a shape association measure. $A_S$ is scale invariant if $S$ is scale invariant.*

Note that if a shape similarity measure $S$ satisfies the conditions of Theorem 2 then it also satisfies the conditions of Theorem 1 hence it can be used for generation of two, generally different, shape association measures $A_S$ by methods given in Theorems 1 and 2.

It is clear that for the reflection invariant shape similarity measure $S$ the methods defined in Theorem 1, 2 will give $A_S(x,y) = 0$ for all time series $x, y$.

## IV. SHAPE ASSOCIATION MEASURES BASED ON MINKOWSKI DISTANCE

To construct translation and scale invariant shape association measures based on Minkowski distance we need to use some transformations of time series that will ensure the fulfillment of these invariance properties. We will consider here standardizations $F$ of time series values that together with Minkowski distance of order $r$ will define dissimilarity measures:

$$D_{r,F}(x,y) = (\sum_{i=1}^{n} |F(x)_i - F(y)_i|^r)^{1/r}.$$

It is clear that $D_{r,F}$ satisfies the properties: $D_{r,F}(x,x) = 0$, $D_{r,F}(x,y) = D_{r,F}(y,x)$. Consider the following possible properties of $F$ formulated for all time series $x,y$ for any real value $q$ and for $p > 0$:

$F(x+q) = F(x)+q$,          (translation additivity)

$F(x+q) = F(x)$,             (translation invariance)

$F(x+y) = F(x)+F(y)$,        (additivity)

$F(px) = pF(x)$,    (scale proportionality or homogeneity)

$F(px) = F(x)$,              (scale invariance)

Note that in literature the translation additivity is often referred to as shift invariance or translation invariance, the scale proportionality is referred to as scale invariance or homogeneity of degree 1. $F$ is said to be **linear** if it is homogeneous and additive. It is clear that from the additivity of $F$ it follows its translation additivity.

A proximity measure $P$ will be referred to as **scale proportional** if

$P(px,y) = pP(x,y)$,      for $p > 0$.

It is clear that the following assertions are fulfilled:

1) $D_{r,F}$ is translation invariant if $F$ is translation invariant;
2) $D_{r,F}$ is scale invariant if $F$ is scale invariant.

Generally we can use different standardizations $F_x$ and $F_y$ applied to time series $x, y$. In this case it will be more correct to say about commutativity $D(x^*,y^*) = D(y^*,x^*)$ with respect to standardized time series $x^* = F_x(x)$, $y^* = F_y(y)$, and instead of shape dissimilarity measure $D_{Fx,Fy}(x,y) = D(F_x(x),F_y(y))$ to say about the shape dissimilarity measure defined on the standardized time series $D(x^*,y^*)$. Also, instead of standardization transformations considered in the following section other types of transformations can be considered.

Suppose $U$ is a strictly decreasing nonnegative function such that $U(0) = 1$ and $S_D(x,y) = U(D_{r,F}(x,y))$ is a similarity measure defined by $U$ and $D_{r,F}$. We can use $S_D(x,y)$ for construction of shape association measure by two methods considered in Theorems 1 and 2 if the properties of sign permutation and weak (or non-) similarity of reflections will be fulfilled for $S_D(x,y)$. The weak similarity property requires $D_{r,F}(x,-x) > 0$ and hence $F(x) \neq F(-x)$, i.e. $F$ should not be an **even** function. Taking this into account we obtain that the sign permutation $D_{r,F}(-x,y) = D_{r,F}(x,-y)$ of the dissimilarity measure $D_{r,F}$ and hence the sign permutation of the similarity measure $S_D(x,y)$ will be fulfilled if standardization $F$ used in Minkowski distance is an **odd function,** i.e. it satisfies:

$F(-x) = -F(x)$.

From Theorem 1 we obtain the following method of construction of time series shape association measure.

***Corollary 2.*** *Suppose $D(x,y) = D_{r,F}(x,y)$ is a dissimilarity measure defined by Minkowski distance, $F$ is a translation invariant odd transformation and $U$ is a strictly decreasing nonnegative function such that $U(0) = 1$, then the function $A_D(x,y)$ defined for all $x,y \neq const$, as follows:*

$$A_D(x,y) = \begin{cases} U(D(x,y)), & if \quad D(x,y) < D(x,-y) \\ -U(D(x,-y)), & if \quad D(x,y) > D(x,-y) \\ 0, & otherwise \end{cases}$$

*is a shape association measure. It is scale invariant if $F$ is scale invariant.*

For example it is possible to use in the formula above one of the following functions:

$$U(D(x,y)) = \frac{K}{D(x,y)+K}, K > 0,$$

$$U(D(x,y)) = \frac{1}{e^{D(x,y)}}.$$

We will say that $F(x)$ satisfies ***r*-normality** if:

$$\sum_{i=1}^{n} |F(x)_i|^r = 1, \quad r = 1,2,....$$

Based on Theorem 2 and the properties of Minkowski distance it can be proposed the following method of construction of time series shape association measures.

***Proposition 1.*** *Suppose $D(x,y) = D_{r,F}(x,y)$ is a dissimilarity measure defined by Minkowski distance, $F$ is a translation invariant odd transformation satisfying r-normality and $W$ is a strictly increasing function such that $W(0) = 0$, $W(2) = 1$, then the function $A_D(x,y)$ defined for all $x,y \neq const$, as follows:*

$A_D(x,y) = W(D(x,-y)) - W(D(x,y))$

*is a shape association measure. $A_D$ is scale invariant if $F$ is scale invariant.*

The simplest functions $W(D(x,y))$ have the form:

$$W(D(x,y)) = \left(\frac{D(x,y)}{2}\right)^p,$$

where $p$ is a positive constant. For $p = 1$ we have from Proposition 1:

$A_D(x,y) = (D(x,-y) - D(x,y))/2$.

For $p = r$ the association measure defined in Proposition 1 has the form:

$$A_D(x,y) = \frac{1}{2^r}\left(\sum_{i=1}^{n}|F(x)_i - F(-y)_i|^r - \sum_{i=1}^{n}|F(x)_i - F(y)_i|^r\right) =$$

$$\frac{1}{2^r}\left(\sum_{i=1}^{n}\left(|F(x)_i + F(y)_i|^r - |F(x)_i - F(y)_i|^r\right)\right).$$

**Corollary 3.** *A shape association measure defined in Proposition 1 coincides with a cosine similarity measure:*

$$A_{cos,F}(x,y) = cos(F(x),F(y)),$$

*if $r=2$, i.e. $D(x,y) = D_{2,F}(x,y)$ is Euclidean distance, and if $W(D)$ is defined as follows:*

$$W(D(x,y)) = \frac{(D(x,y))^2}{4}.$$

## V. STANDARDIZATION OF TIME SERIES VALUES

A transformation $F$ of time series $x$ of length $n$ into time series $F(x)$ of the same length is said to be a **standardization** if for all non-constant time series $x$ it is fulfilled:

$F(F(x)) = F(x).$   (idempotency)

Two additional requirements on standardization transformation can be considered:

$F(x) \neq const$   if $x \neq const$,

$F(q_{(n)}) = 0_{(n)}$, for any real value $q$.

A time series $x$ is said to be **in a standard form wrt a standardization $F$** if $F(x) = x$. As it follows from the definitions, a standardization transforms any time series $x$ into a standard form $F(x)$.

A transformation $E$ of time series $x$ of length $n$ into real value $E(x)$ is said to be an **estimate** of $x$. We can denote $F=E$ and introduce for estimates the properties of translation additivity, translation invariance etc, discussed in the previous section.

**Proposition 2.** *Suppose $E_1(x)$ is a translation additive estimate such that $E(q_{(n)})=q$, then the transformation*

$F(x) = x - E(x),$

*is a translation invariant standardization such that*

$E(F(x)) = 0.$

*If $E(x)$ is an odd function, then $F(x)$ is an odd function. If $E(x)$ is scale proportional then $F(x)$ is scale proportional.*

**Proposition 3.** *Suppose $E_1(x)$ is a translation additive and scale proportional estimate such that $E(q_{(n)})=q$, and $E_2(x) \neq 0$ is a translation invariant and scale proportional estimate then the transformation*

$F(x) = (x - E_1(x))/E_2(x)$

*is a translation invariant and scale invariant standardization such that*

$E_1(F(x)) = 0.$

*If $E_1(x)$ is an odd function and $E_2(x)$ is an even function, then $F(x)$ is an odd function.*

*If $E_2(x) = (\sum_{i=1}^{n}|x_i - E_1(x)_i|^r)^{1/r}$ then $F(x)$ satisfies r-normality.*

Consider time series $x = (x_1,\ldots,x_n)$. Reorder its values in an ascending order and denote $x_{(k)}$ the $k$-th lowest value of $x$, i.e.:

$x_{(1)} \leq \ldots \leq x_{(k)} \leq \ldots \leq x_{(n)}.$

An estimate $E$ is said to be a **mean** if it satisfies the condition [14]:

$\min\{x_1,\ldots,x_n\} \leq E(x) \leq \max\{x_1,\ldots,x_n\}.$

Note that $E(q_{(n)}) = q$. Consider some means of $x$ [14, 18, 19, 23, 25-27]:

$MIN(x) = \min\{x_1,\ldots,x_n\},$   (minimum)

$MAX(x) = \max\{x_1,\ldots,x_n\},$   (maximum)

$MDR(x) = (MIN(x) + MAX(x))/2,$   (midrange)

$PR_k(x) = x_k,$   $k = 1,\ldots,n,$   (projection)

$OS_k(x) = x_{(k)},$   $k = 1,\ldots,n,$   (order statistics)

where $OS_1(x) = MIN(x)$, $OS_n(x) = MAX(x)$,

$MED(x) = x_{((n+1)/2)},$   for odd $n$,   (median)

$MED(x) = (x_{(n/2)} + x_{(n/2+1)})/2,$   for even $n$,   (median)

$TM_m(x) = \frac{1}{n-2m}\sum_{j=m+1}^{n-m} x_{(j)},$   $m < n/2,$   (truncated mean)

$GMDR_{k,m}(x) = \frac{1}{2(m-k)}\left(\sum_{j=k+1}^{m} x_{(j)} + \sum_{j=n-m+1}^{n-k} x_{(j)}\right),$

where $0 \leq k < m < n/2,$   (generalized midrange)

$AM(x) = \bar{x} = \frac{1}{n}\sum_{j=1}^{n} x_j,$   (arithmetic mean)

$WAM_w(x) = \sum_{j=1}^{n} w_j x_j$   (weighted arithmetic mean)

$OWA_w(x) = \sum_{j=1}^{n} w_j x_{(j)},$ (ordered weighted arithmetic mean)

where $w = (w_1,\ldots,w_n)$, are nonnegative weights such that $\sum_{j=1}^{n} w_j = 1.$

The midrange can be obtained from $GMDR_{k,m}$ if $k=0$, $m=1$. Note that an aggregation of a truncated mean and a generalized midrange for $k=0$ gives an arithmetic mean as follows: $AM(x) = ((n-2m)*TM_m(x) + 2m*GMDR_{0,m}(x))/n.$

The means are often referred to as location measures, measures of the center or central tendency, an estimate $E_2$ is referred to as a scale measure, the standardization $F(x) = x - E(x)$ is referred to as centering. The standardization methods given in Propositions 2 and 3 are widely used in practice but we do not know definitions of neither standardization in general form as an idempotent transformation no generalized midrange.

It is clear that all means considered above are translation additive and scale proportional means. They can be used as estimates $E$ and $E_1$ in construction of standardizations given by Propositions 2 and 3. It can be shown that the following means are **odd functions**: $MDR$, $PR_k$, $MED$, $TM_m$, $GMRD_{k,m}$, $AM$, $WAM_w$. These estimates can be used in the methods of construction of shape association measures considered above. As an estimate $E_2$ in Proposition 3 one can use a **range**:

$$R(x) = MAX(x) - MIN(x),$$

which is translation invariant and scale proportional. Below are some examples of standardizations:

$$F_1(x)_i = x_i - \bar{x},$$

$$F_2(x)_i = x_i - MIN(x),$$

$$F_3(x)_i = \frac{x_i - \bar{x}}{\sqrt{\sum_{j=1}^{n}(x_j - \bar{x})^2}},$$

$$F_4(x)_i = \frac{x_i - GMDR_{k,m}}{\sqrt[r]{\sum_{j=1}^{n}|x_j - GMDR_{k,m}|^r}}.$$

$F_1$ and $F_2$ are translation invariant and scale proportional, $F_3$ and $F_4$ are translation and scale invariant satisfying 2-normality and $r$-normality respectively, $F_1$, $F_3$ and $F_4$ are odd functions.

## VI. SOME EXAMPLES

From Proposition 1 and Corollary 3 it follows

***Corollary* 4.** *A shape association measure*

$$A_D(x,y) = W(D(x,-y)) - W(D(x,y))$$

*defined in Proposition* 1 *by Euclidean distance ($r=2$), $W$ function*

$$W(D(x,y)) = \frac{(D(x,y))^2}{4},$$

*and standardization $F_3$ defined above coincides with the sample Pearson's correlation coefficient:*

$$A(x,y) = \frac{\sum_{i=1}^{n}(x_i - \bar{x})(y_i - \bar{y})}{\sqrt{\sum_{i=1}^{n}(x_i - \bar{x})^2 \cdot \sum_{i=1}^{n}(y_i - \bar{y})^2}}.$$

Considered methods of time series standardization and construction of shape association measures give possibility to construct dozens of time series shape association measures that can be used for analysis of direct and inverse relationships between time series. For example, the dissimilarity measure

$$D(x,y) = (\sum_{i=1}^{n}|F_4(x)_i - F_4(y)_i|^r)^{1/r},$$

defined by Minkowski distance and standardization $F_4$ can be used for generation time series shape association measures by means of functions $U(D(x,y)) = e^{-D(x,y)}$ or $U(D(x,y)) = 1/(1+D(x,y))$ applying the method described in Corollary 2 or by means of function $W(D(x,y)) = \left(\frac{D(x,y)}{2}\right)^p$ by method described in Proposition 1. In the simplest case we can consider the following shape association measure:

$$A(x,y) = \frac{\sum_{i=1}^{n}(x_i - GMDR_{k,m}(x))(y_i - GMDR_{k,m}(y))}{\sqrt{\sum_{i=1}^{n}(x_i - GMDR_{k,m}(x))^2 \cdot \sum_{i=1}^{n}(y_i - GMDR_{k,m}(x))^2}},$$

that is similar to Person's correlation coefficient but uses another central measure. In [5], it was shown that the correlation coefficient is not relevant for measuring shape association of time series. So, before applying a constructed shape association measure on real tasks it is desirable to test its efficiency on testing examples.

## VII. SOME RESULTS OF ANALYSIS OF PERFORMANCE OF CONSIDERED METHODS

The goal of the paper was to propose a normative approach to analysis and construction of shape association measures from the point of view of some reasonable requirements on the shape association measures that can be generated. A dozen of association measures can be constructed based on the methods described in the paper. As it follows from the construction, these measures satisfy translation and/or scale invariance. Comparative analysis of these measures is a theme of a special investigation. Nevertheless the performance analysis of some of them on a benchmark example will be useful.

As a benchmark example for a comparative analysis of time series shape association measures we used data from file RealityCheck.dat from UCR time series data mining archive [15]. Time series from this file are presented in Fig. 1. This example is discussed also in [6, 28]. The clustering of these time series proposed by owners of this data is given in Fig. 2, see also [24]. It is surprising that they could not discover that the time series 9 and 10 are almost inverse to 1 and 4 time series and due to high relationship between them should be clustered in one cluster. The results of association analysis of data from this example based on local trend association measure considered in [6] are shown in Fig. 3, where two objects are connected by edge with the weight equal to the shape association value between corresponding time series. Fig. 3 shows only the weights greater than 0.70 by absolute value. Any reasonable clustering algorithm based on similarity function $S_A(x,y) = abs(A(x,y))$ obtained for this example in [6] by local trend association measure will construct 5 clusters presented in Fig. 3 by connected subgraphs, and hence will join objects {1,4,9,10} in one cluster. We will consider these clusters as true clusters. Note that these clusters are presented also in dendrogram reported in [28] and obtained by independent method using SOM with global characteristic measures.

In this work it was done the comparative analysis of shape association measures $A_D$ constructed by means of Euclidean distance ($r=2$) and standardizations $MDR$, $PR_k$,($k=2$), $MED$, $TM_m$, ($m=2$), $GMRD_{k,m}$, ($k=0$, $m=2$), $AM$ in two methods considered in: 1) Corollary 2; and 2) Proposition 1, Corollary 3. In method given in Corollary 2 the following function $U$ was applied:

$$U(D(x,y)) = \frac{1}{D(x,y)+1}.$$

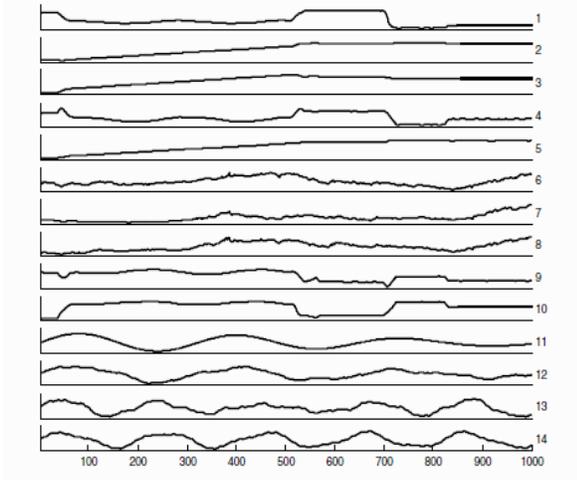

Fig. 1. Time series from the file RealityCheck.dat

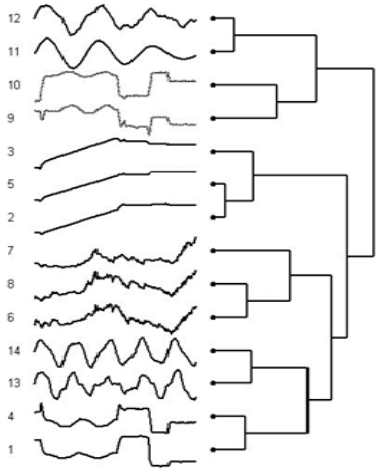

Fig. 2. Results of clustering proposed by owners of RaealityCheck.dat

The single linkage clustering algorithm was applied to all similarity measures $S_A(x,y)=abs(A_D(x,y))$ obtained from considered association measures $A_D$. For association and similarity measures obtained by both methods from standardizations $MDR$, $MED$, $GMRD_{k,m}$, ($k=0$, $m=2$), the clustering algorithm has constructed dendrograms containing true clusters {1,4,9,10}, {2,3,5}, {6,7,8}, {11,12}, {13,14}. For association and similarity measures obtained from standardizations $PR_k$,($k=2$), $TM_m$, ($m=2$), $AM$ the dendrograms constructed by clustering algorithm do not contained all true clusters.

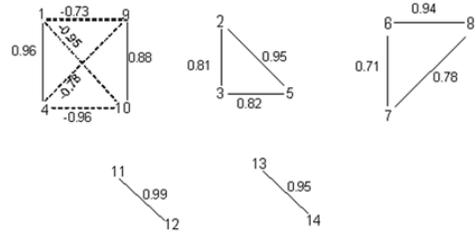

Fig. 3. Association graph of time series from RealityCheck.dat obtained by local trend association measure in [6]. Only high positive (solid lines) and negative (dotted lines) association values are presented.

VIII. DISCUSSION AND CONCLUSIONS

The properties of translation invariance (or additivity) and scale invariance (homogeneity) of time series similarity measures together with normalization of time series have been discussed in many works [1, 2, 10, 12, 13, 29, 30], see also [20]. In [30] the reverse time series, obtained by multiplying original time series by -1, have been also considered. But to the best of our knowledge the properties of time series shape association measures have been systematically studied only in the pioneering work [9], see also [5].

In the present paper the theoretical foundations of the methods of construction of these measures are proposed and studied. These methods are considered without references on possible statistical properties of time series. Such approach is more traditional for time series data mining and time series clustering. But of course it would be interesting to analyze statistical properties of the new association measures in the presence of errors in time series values. Generally, proposed measures can be used also for measuring associations between variables instead of or complimentary to Pearson' correlation coefficient. An influence of standardization methods on results of cluster analysis has been studied in several works [18, 19, 25-27] but sometimes the conclusions of such analysis are contradictory. Description and applications of local trend association measures based on moving approximation transform can be found in [6, 3, 5]. In comparison with similarity measures the association measures give more powerful instrument for analysis of possible relationships between time series. Such analysis can be useful in data driven analysis of complex systems when a dynamics of system elements is given by the sets of time series. A mathematical analysis of complex systems in economics, finance, petroleum industry, meteorology etc. is usually based on simulation and modeling of system dynamics but often such modeling is cost prohibitive or cannot give adequate results due to systems complexity or absence of information necessary for models. In such cases a data driven approach to analysis of complex systems including analysis of relationships between dynamics of system elements can be helpful and complementary to conventional modeling [3, 5, 6, 8, 17, 21, 22].

ACKNOWLEDGMENT

The presented results have been partially supported by the projects D.00507 of IMP and by PIMAyC Research Program of IMP.